%% file: main.tex
\documentclass[letterpaper,12pt]{extarticle}

\usepackage{import}
\usepackage{local}
\usepackage[left=.75in,top=.75in,bottom=.75in,right=.75in]{geometry}

\usepackage[utf8]{inputenc}
\usepackage[T1]{fontenc}

\usepackage{float}
\usepackage{physics}
\usepackage{graphicx}
\usepackage{dcolumn}
\usepackage{bm}
\usepackage{mathrsfs}
\usepackage{amssymb}
\usepackage{amsmath}
\usepackage{enumitem}
\usepackage[normalem]{ulem}
\usepackage{wrapfig}
\usepackage{csquotes}
\usepackage{etoc}
\usepackage[ruled,vlined]{algorithm2e}
\usepackage{booktabs}

\usepackage{lineno}

\usepackage{caption}
\usepackage{url}
\usepackage{hyperref}
\usepackage{bookmark}

\hypersetup{%
  breaklinks=true,
  colorlinks=true,
  linkcolor=black,
  filecolor=black,
  urlcolor=black,
  citecolor=black,
  pdfborder=0 0 0,
}

\usepackage[
    backend=biber,
    bibstyle=trad-unsrt,
    citestyle=numeric-comp,
    sorting=none,
    isbn=false,
    eprint=false,
    maxnames=500,
    url=true,
    doi=true,
    eprint=false,
]{biblatex}
\renewbibmacro{in:}{}
\AtEveryBibitem{
    \clearlist{language}
}
\AtEveryBibitem{\clearfield{month}}
\usepackage{subfiles}
\usepackage{outlines}

\usepackage{xcolor}

\newcommand{\citep}[1]{\cite{#1}}

\begin{document}

\title{\large Learning to model pediatric asthma exacerbation from multiple risk factors: a case study in coastal Virginia}
\author{\textbf{Jonathan Colen\textcolor{Accent}{\textsuperscript{1,2}$\dagger$}, 
        Eric Werner\textcolor{Accent}{\textsuperscript{3,4,5}},
        Maryam Golbazi\textcolor{Accent}{\textsuperscript{1,2,6}},
        Heather Richter\textcolor{Accent}{\textsuperscript{1}},
        Diana McSpadden\textcolor{Accent}{\textsuperscript{1,7}},
        Amy Quinn\textcolor{Accent}{\textsuperscript{4}},
        Jocel Santos\textcolor{Accent}{\textsuperscript{4}},
        Mary Jane Darling\textcolor{Accent}{\textsuperscript{4}},
        Mary Margaret Gleason\textcolor{Accent}{\textsuperscript{3,8}}
} \\
\begin{small}
    \textcolor{Accent}{\textsuperscript{1}} 
    Joint Institute on Advanced Computing for Energy and Science, Old Dominion University, Suffolk, VA, 23535, U.S.A. \\
    \textcolor{Accent}{\textsuperscript{2}} 
    School of Data Science, Old Dominion University, Norfolk, VA, 23529, U.S.A. \\
    \textcolor{Accent}{\textsuperscript{3}}
    Eastern Virginia Medical School, Macon \& Joan Brock Virginia Health Sciences, Old Dominion University, Norfolk, VA, 23501, U.S.A. \\
    \textcolor{Accent}{\textsuperscript{4}} 
    Children's Hospital of the King's Daughters, Norfolk, VA, 23507, U.S.A. \\
    \textcolor{Accent}{\textsuperscript{5}} 
    Children's Specialty Group, Norfolk, VA, 23507, U.S.A. \\
    \textcolor{Accent}{\textsuperscript{6}}
    Institute for Coastal Adaptation and Resilience, Old Dominion University, Norfolk, VA, 23508, U.S.A. \\
    \textcolor{Accent}{\textsuperscript{7}} 
    Chief Data Office, Thomas Jefferson National Accelerator Facility, Newport News, VA, 23606, U.S.A. \\
    \textcolor{Accent}{\textsuperscript{8}} 
    Department of Psychiatry, Boston Children's Hospital, Boston, MA, 02115, U.S.A. \\
    \textcolor{Accent}{\textsuperscript{$\dagger$}} Correspondence: \textcolor{Accent}{jcolen@odu.edu} \\
\end{small}
}

\maketitle

\section*{Abstract}
\input{chapters/abstract}
\newpage

\input{chapters/introduction}

\input{chapters/methods}

\input{chapters/results}

\input{chapters/discussion}

\input{chapters/limitations}

\printbibliography

\section{Statements and Declarations}

\subsection{Funding}

JC, MG, and DS acknowledge support from the Hampton Roads Biomedical Research Consortium as part of the efforts associated with the Joint Institute for Advanced Computing on Energy and Science between Old Dominion University and Thomas Jefferson National Accelerator Facility. 
This research was supported by the Research Computing Clusters at Old Dominion University. 
DS acknowledges support from SURATech, LLC, operating the Thomas Jefferson National Accelerator Facility for the U.S. Department of Energy under Contract No. 89243126CSC000213.

\subsection{Competing Interests}

The authors declare they have no financial interests.

\subsection{Author Contributions}

JC, EW, HR, DS, and MMG conceptualized the study.
JC, DS, AQ, JS, and MJD curated data.
JC developed models and conducted analyses.
JC, EW, MG, and MMG interpreted results.
JC and MG wrote the original draft.
All authors contributed to reviewing and editing the manuscript.

\subsection{Ethics approval}

This research was approved by the Institutional Review Board for Macon \& Joan Brock Virginia Health Sciences at Old Dominion University (IRB Protocol No. 23-10-WC-0267).

\subsection{Data Availability}

Modeling and analysis code is available online at \url{https://github.com/jcolen/pediatric_asthma}

\appendix
\newpage

\renewcommand{\thefigure}{S\arabic{figure}}
\renewcommand{\thetable}{S\arabic{table}}
\renewcommand{\thealgocf}{S\arabic{algocf}}
\renewcommand{\theequation}{S\arabic{equation}}

\setcounter{equation}{0}
\setcounter{figure}{0}
\setcounter{algocf}{0}
\setcounter{table}{0}

\section{\large Supplementary Information}
\input{chapters/appendix}

\end{document}

%% file: chapters/abstract.tex
Childhood asthma is a common illness exacerbated by air pollution as well as meteorological and neighborhood-level socioeconomic factors. 
Modeling asthma exacerbation (AE) in large spatiotemporal datasets requires disentangling impacts from multiple contributors.
In this case study, we compared three techniques that balance predictive power with interpretability to predict AE in Hampton Roads, a coastal Virginia region comprising 7 cities and over 1.5 million people.
After collating ambient air pollution measurements, weather data, and measures of neighborhood opportunity, we modeled zip code-level acute AE visits to a regional children's hospital and affiliated providers from 2018-2023. 
Generalized linear models (GLM) provided a baseline while neural networks (NN) served as a maximally predictive target.
To bridge between statistical models and deep learning, we developed a framework based on sparse dictionary learning to identify and interpret parsimonious nonlinear interacting equations.
After comparing each model's predictive performance, we estimated relative risks for AE due to input exposure variables and found consensus across frameworks.
Our work links statistical and interpretable machine learning models to highlight possible synergistic interactions influencing AE, and may enable future studies to  guide public health interventions in coastal Virginia.

%% file: chapters/introduction.tex
\section{Introduction}

Asthma is a common childhood illness affecting nearly 7\% of children in the United States~\citep{cdc_asthma_2024,akinbami_trends_2012,akinbami_changing_2016}.
Ambient air pollution has been associated with both the development and exacerbation of asthma~\citep{anderson2012long,kelly2011air,zhou2024effect,milligan_asthma_2016}.
Fine particulate matter (PM$_{2.5}$), nitrogen dioxide (NO$_2$), ozone (O$_3$), and sulfur dioxide (SO$_2$) have been consistently identified as key pollutants influencing asthma rates~\citep{epa2025,who2021,guarnieri2014outdoor,kiser_particulate_2020,gauderman2004,gauderman2015association}. 
Additional risks include meteorological factors such as temperature~\citep{schinasi_associations_2022} and precipitation~\citep{soneja_exposure_2016} as well as viral respiratory infections~\citep{busse_role_2010} and socioeconomic factors such as neighborhood poverty~\citep{keet_urban_2017,navanandan_predicting_2021}.
Assessing collective impacts therefore requires accounting for  the timing, localization, and potential interactions of multiple contributors~\citep{hsu_differential_2020,shmool_spatio-temporal_2016,huang_prediction_2025}. 

Air pollutants may be emitted directly into the atmosphere as primary pollutants or form from atmospheric chemical reactions as secondary pollutants~\citep{naaqs}. 
Because atmospheric processes such as chemical transformation, transport, and diffusion are highly dynamic, pollutant concentrations do not always peak near their sources. 
Emissions from distant wildfires can drift over long distances, where they may degrade air quality and influence asthma-related health outcomes~\citep{wang_differential_2025,reid_differential_2016}. 
NO$_2$, widely used as an indicator of traffic-related air pollution, is associated with degraded air quality conditions that can contribute to respiratory irritation and increased sensitivity \citep{chen2022health,khreis2017exposure,guarnieri2014outdoor}. 
These impacts are linked to pollutant-driven chemical processes, including the formation of reactive oxidants and interactions with atmospheric constituents, which can influence respiratory function and increase susceptibility to asthma-related outcomes~\citep{kelly2011air,hiltermann1998asthma, zhou2024effect}. 

Exposure timing, pollutant mixtures, and population susceptibility modulate the relationship between air pollution and asthma~\citep{gehring_exposure_2015,oneill2003health}. 
Early-life exposure during prenatal and early childhood periods has been shown to increase the risk of asthma development~\citep{gehring_exposure_2015}. 
Socioeconomically disadvantaged populations and urban communities often experience disproportionately higher exposure to traffic-related pollutants and environmental stressors such as extreme heat~\citep{golbazi2025high}.
This may promote conditions that increase air pollution levels, particularly O$_3$, contributing to disparities in asthma prevalence and severity~\citep{oneill2003health}. 
In addition, emerging research suggests synergistic interactions between air pollutants and respiratory infections, which can amplify airway inflammation and trigger asthma exacerbations~\citep{damato2016effects}.
While experimental and epidemiological evidence support causal relationships, uncertainties remain regarding pollutant-specific toxicity and threshold effects~\citep{who2021}.

New modeling approaches may capture impacts and interactions from multiple environmental factors.
Although studies have examined individual interactions between chosen pollutants and aeroallergens~\citep{cakmak_does_2012} or wildfire smoke~\citep{wang_differential_2025}, incorporating interactions within standard statistical models may not improve risk prediction~\citep{huang_prediction_2025}.
More recently, deep learning~\citep{goodfellow_deep_2016,lecun_deep_2015} has shown promise in predicting asthma-related hospital and emergency visits~\citep{hwang_prediction_2023,lopez_deep_2023,alsaad_predicting_2022}.
Trained on data, these tools capture complex interactions and nonlinear features to achieve predictive power~\citep{goodfellow_deep_2016}.
While these models are predictive, they are not explanatory. 
\textit{Post hoc} analysis, which views trained neural networks as a target for subsequent investigation and simplification~\citep{murdoch_definitions_2019}, has emerged as a promising path to discover interpretable mathematical rules in biophysics and biology~\citep{soelistyo_learning_2022,achar_universal_2022,schmitt_machine_2024,colen_interpreting_2024}.
Providing transparent justifications for black-box model outputs may also alleviate concerns regarding trustworthy applications of neural networks in health domains~\citep{abgrall_should_2024,chaddad_survey_2023}.

To study the trade-off between interpretability and predictive power, we demonstrated and compared three techniques to model childhood asthma exacerbation (AE) in Hampton Roads, Virginia.
Using ambient air pollution measurements, meteorological data, and neighborhood-level socioeconomic factors, we predicted daily visit counts to a regional children's hospital and affiliated providers for acute AE from 2018-2023.
Generalized linear models (GLM)~\citep{nelder_generalized_1972,mccullagh_generalized_2019} provided a predictive baseline and straightforward interpretation via relative risks~\citep{naimi_estimating_2020,zou_modified_2004}. 
Next, we applied neural networks (NN) as a ``maximal model" and compared feature saliency measurements to GLM coefficients. 
Finally, we developed a framework that couples Poisson regression with sparse dictionary modeling~\citep{brunton_discovering_2016,kaptanoglu_pysindy_2022} to learn parsimonious nonlinear interacting equations that bridge the gap between minimal and maximal models.
Analyzing correspondences between GLM, NN, and sparse models within a shared interpretive framework opens a window into the black box and sheds light on environmental factors and interactions that collectively exacerbate asthma in coastal Virginia. 

%% file: chapters/methods.tex
\section{Materials and Methods}

\subsection{Study data}

\begin{wrapfigure}[23]{r}{0.5\textwidth}
    \vspace{-2em}
    \centering
    \includegraphics[width=\linewidth]{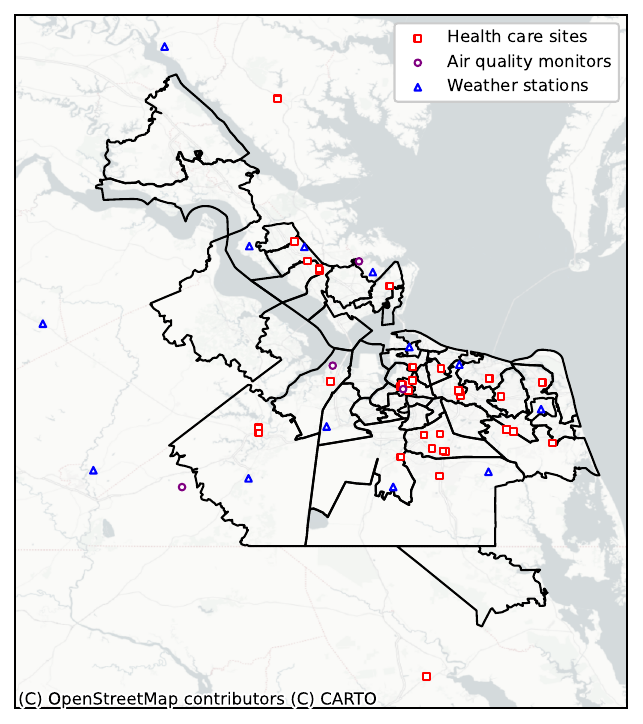}
    \caption{Hampton Roads study area showing health care sites (red), AQS air monitor locations (purple), and weather stations (blue).}
    \label{fig:study_area}
\end{wrapfigure}\leavevmode
The IRB-approved study used a de-identified dataset containing visits from patients aged 0-17 at Children's Hospital of the King's Daughters (CHKD) locations from May 1, 2018 through July 31, 2023. The original dataset included 179,021 visits with ICD10 codes R05 (cough), J06 (accute upper respiratory infection), and J45 (asthma). For this study, we used only visits with the most precise asthma-specific code (J45, N=13,750). We additionally selected only zip codes containing at least 100 total visits over the study period (42 zip codes, N=10,250) The total study area, relevant zip code tabulation areas (ZCTA), and included healthcare sites are shown in Figure~\ref{fig:study_area}. 
As our study focuses on respiratory illness, we also collected the number of daily positive COVID 19 tests in Virginia during the study period from HealthData.gov~\citep{cdc_covid-19_nodate}. 

\subsubsection{Environmental data}

We collected air quality measurements during the study period from federal reference monitors accessed through the EPA Air Quality System (AQS)~\citep{us_epa_air_2013}. The locations of the three available monitors are shown in Figure~\ref{fig:study_area}. We collected NO$_2$, SO$_2$, CO, PM$_{2.5}$ and PM$_{10}$ and interpolated values to the centroids of each ZCTA using radial basis function (RBF) interpolation. We did not use O$_3$ in this study because our local reference monitors did not measure O$_3$ during winter months. 

For each zip code tabulation area, we collected meteorological data from the NOAA Local Climatological Service Data v2~\citep{noaa_lcdv2_2023} including daily total precipitation, relative humidity, and average dry bulb temperature. The locations of the thirteen weather stations are shown in Figure~\ref{fig:study_area}.
To support follow-up analysis, atmospheric smoke density measurements were collected from the NOAA Hazard Mapping System (HMS)~\citep{noaa_hms_nodate}. Using an area-weighted average, we assigned a daily atmospheric smoke level to each ZCTA.

\subsubsection{Socioeconomic data}

To account for socioeconomic influences, we collected zip code-level Child Opportunity Index 3.0 (COI) data for each year of the study period.
COI is a composite metric of neighborhood conditions associated with child health and well-being calculated using 44 indicators across three domains: education, health and environment, and social and economic~\citep{coi_child_nodate}.
We used nationally-normed COI scores, which take values from 1 to 100. 
COI also estimates the local child population (age 0-17) by combining decennial Census and American Community Survey data~\citep{coi_child_nodate}.
We used this as a population offset within each ZCTA.
To evaluate model sensitivity, we also collected yearly zip code-level Social Vulnerability Index (SVI) data. 
SVI combines 16 variables from the 5-year American Community Survey across four domains: socioeconomic status, household characteristics, racial \& ethnic minority status, and housing type \& transportation~\citep{cdc_social_2024}.
We collected nationwide SVI percentiles for 2022, the only year when zip code-level measurements were available.

\subsection{Predictive models}

\begin{figure}[t]
    \centering
    \includegraphics[width=0.7\linewidth]{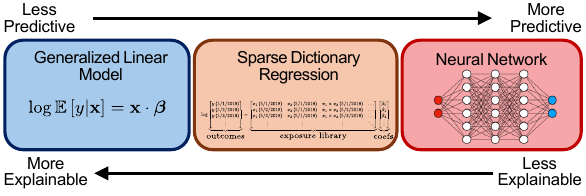}
    \caption{Modeling methods range from less predictive but more explainable (GLM) to more predictive but less explainable (NN). Sparse dictionary regression bridges the gap between two techniques.}
    \label{fig:workflow}
\end{figure}

We compared three modeling methods to predict daily AE and infer associations between exposures and outcomes. 
These approaches are shown schematically in Figure~\ref{fig:workflow}.
Generalized linear models (GLM)~\citep{nelder_generalized_1972,mccullagh_generalized_2019} make transparent predictions and support straightforward interpretation.
Neural networks~\citep{goodfellow_deep_2016,lecun_deep_2015} leverage nonlinear features to achieve greater predictive power, but are black boxes.
We introduce a middle ground approach that couples generalized linear modeling with nonlinear sparse dictionary learning, which selects parsimonious predictive equations from a library of candidate functions~\citep{brunton_discovering_2016}. 
This technique produces interpretable mathematical equations that leverage a minimal number of complex nonlinear features.

The models predicted the daily number of acute AE visits in patients from each ZCTA. 
We evaluated each model using 5 trials of 3-fold cross validation. 
Zip code was used as a grouping variable to eliminate geographic overlap between train and test sets.
After training, each model predicted AE visits for its respective test ZCTAs over the entire study period.
We evaluated model accuracy using test $R^2$ score and mean absolute error (MAE) for daily AE visits aggregated over all trials and folds, representing average predictive performance over the entire study area. 
To interpret model predictions, we computed and compared relative risks (RR) for each model~\citep{naimi_estimating_2020,zou_modified_2004}. 
Following previous work~\citep{reid_differential_2016}, we scaled each continuous exposure by its interquartile range, thus the effect represents the relative change associated with a 1-IQR increase in each parameter (see Table~\ref{tab:descriptive} for percentiles and quartiles).
We computed average RR and standard deviations by propagating uncertainty across all trials and folds. 
To assess model sensitivity, we repeated each analysis substituting COI for SVI. 
To examine how factors and interactions may influence AE over time~\citep{hsu_differential_2020,shmool_spatio-temporal_2016,huang_prediction_2025}, we considered time lags between exposures and outcomes of 0 days, 1 day, 3 days, 5 days, and 7 days.
We chose this range considering results on how air pollutant exposure lags of up to 6 days may modulate AE~\citep{orellano_effect_2017}.
Code for all models is available online~\citep{jcolen_github}.

\subsubsection{Generalized linear models}

We used a quasi-Poisson generalized linear models (GLM)~\citep{nelder_generalized_1972,mccullagh_generalized_2019} with a log-link function to model ZCTA-level daily visits with ICD10 code J45 (asthma). 
The models included linear coefficients for air quality measures (NO$_2$, SO$_2$, CO, PM$_{2.5}$, PM$_{10}$), weather (temperature, relative humidity, precipitation), and COI overall percentile. 
We included a log population offset to each zip code (ages 0-17) and one-hot encoded indicator variables for day of week and month of year. 
We included an additional variable containing $\log 1 + C$ where $C$ is the daily number of positive covid tests. 
The model is captured by the following equation:

\begin{align}
\log \mathbb{E} [ y | \vec{X} ] &= 
\left\{ \text{NO}_2, \text{SO}_2, \text{CO}, \text{PM}_{2.5}, \text{PM}_{10}\right\} \cdot \vec{\beta}_{AQ} & \mbox{Air quality} \notag \\
&+ \ \text{Temp} \cdot \beta_{T} + \text{RelHumid} \cdot \beta_{RH} + \text{TotalPrecip} \cdot \beta_{P}& \mbox {Weather} \notag \\ 
&+ \ \text{COI} \cdot \beta_{COI} & \mbox {Child Opportunity} \notag \\
&+ \ \left\{ \text{Sun}, \text{Mon}, \text{Tue}, \dots, \text{Sat}\right\} \cdot \vec{\beta}_{day} & \mbox{Weekday indicator} \notag \\
&+ \ \left\{  \text{Jan}, \text{Feb}, \text{Mar}, \dots, \text{Dec}\right\} \cdot \vec{\beta}_{month} & \mbox{Month indicator} \notag \\
&+ \ \beta_C \log{C} + \log{P_{0-17}} + \alpha & \mbox{Infections and Population}
\label{eq:gam}
\end{align}

\noindent
We implemented GLMs in Python using the statsmodels package~\citep{seabold2010statsmodels}. 

\subsubsection{Neural networks}

To predict patient visits using a higher-order nonlinear function, we trained a neural network (NN) with 2 hidden layers each containing 256 neurons with hyperbolic tangent activations and a linear output layer. 
The NN received an input vector of each independent variable listed in (\ref{eq:gam}) and predicted $\log y - \log P_{0-17}$, or the log-rate of patient visits from each ZCTA. This mirrors the log population offset used in the GAM. The network output $f_{NN}$ and total visit predictions $y$ are related by the following equation:
\begin{align}
    y = \exp \left[ f_{NN}(\vec{X}) + \log P_{0-17} \right]
    \label{eq:nn_predict}
\end{align}
We implemented neural networks using PyTorch~\citep{paszke_pytorch_2019} and trained for 100 epochs using an Adam optimizer (learning rate 0.0003, batch size 256) to minimize a Poisson-distributed negative log-likelihood objective function.
To measure relative risk in the neural network, we used averaged saliency over the dataset. Saliency is a measure of the linear response of the neural network to a change in its inputs and is computed via automatic gradient backpropagation~\citep{simonyan_deep_2014,adebayo_sanity_2018}. We approximated the risk ratio by aggregating neural network saliency over the test dataset as follows. 
\begin{align}
    RR^{NN}_i = \exp \left[ \frac{1}{N} \sum_{n=1}^N \frac{\partial f_{NN} (\vec{X}_n) }{ \partial X_{n,i}} \right]
    \label{eq:nn_rr}
\end{align}
For a single-layer linear NN, Eq.~\ref{eq:nn_predict} reduces to a GLM prediction with coefficients equal to the network weights. In this limit, Eq.~\ref{eq:nn_rr} recovers the GLM expression $RR_i = \exp \beta_i$. 

\subsubsection{Sparse dictionary regression}

To perform sparse model identification, we made two modifications to the GLM fitting procedure.
First, we constructed a feature library containing all second order polynomial terms containing continuous variables in (\ref{eq:gam}). The feature library $F$ also contained the weekday and monthly indicators from (\ref{eq:gam}), see the following definitions:
\begin{align}
    P(\vec{X}) &= \left\{ \text{NO}_2, \text{SO}_2, \text{CO}, \text{PM}_{2.5}, \text{PM}_{10}, \text{Temp}, \text{RelHumid}, \text{TotalPrecip}, \text{SVI}, \log \text{C} \right\} \\
    F(\vec{X}) &= \left\{ 1, x, x^2, x\cdot y \ | \ x, y \in P(\vec{X}) \right\}  \cup \left\{ \text{Sun}, \text{Mon}, \dots, \text{Jan}, \text{Feb}, \dots \right\}
    \label{eq:library}
\end{align}
Next, we selected a minimal subset of terms in $F$ using sequentially-thresholded least squares (STLSQ) optimization~\citep{brunton_discovering_2016}. 
This algorithm, originally developed for dynamical systems, performs multiple iterations of model-fitting~\citep{brunton_discovering_2016}. 
At each step, the algorithm removes library terms whose fitted coefficients are below a preset threshold $\tau$.
While the original STLSQ algorithm used linear (ridge) regression, here we performed quasi-Poisson regression. At each optimization step, the model aimed to find coefficients $\vec{w}$ to minimize the residual $ \log y - \log P_{0-17} - F \cdot \vec{w}$. 
The complete algorithm is shown in Alg.~\ref{alg:spr}. 
This method reproduces a standard GLM model using $\tau = 0$ and a linear library $F(\vec{X}) = P(\vec{X}) \cup \left\{ \text{Sun}, \text{Mon}, \dots, \text{Jan}, \text{Feb}, \dots \right\}$. 
Thus, we could compare consistently against the GLM and NN models. 
To evaluate tradeoffs between model complexity and performance, we varied our threshold parameter $\tau$ and compared model performance for 25 values in the range $\left[0.01, 50\right]$. 
We computed relative risks and standard deviations for the sparse models by propagating coefficient uncertainties as described in the Appendix. 
We implemented sparse dictionary regression models in Python by extending the GLM module in the \texttt{statsmodels} package~\citep{seabold2010statsmodels}.

The feature library also includes quadratic terms (\ref{eq:library}) representing interactions between exposures. 
The RERI for two exposures $X_1, X_2$ is~\citep{correia_estimating_2018,richardson_estimation_2009}
\begin{align}
    RERI_{12} = RR_{12} - RR_{1} - RR_{2} + 1
\end{align}
We computed the relative excess risk due to interaction (RERI) for each model using the associated coefficients for each term. 
We aggregated the RERI across all trials to estimate the additive effects of interactions selected by the optimization procedure.

%% file: chapters/results.tex
\begin{table}[h]
    \centering
\begin{tabular}{lrrrrrrr}
\toprule
 & p5 & p10 & p25 & p50 & p75 & p90 & p95 \\
\midrule
\textbf{Ambient air quality} \\
NO$_2$ (ppb) & 1.99 & 2.54 & 3.53 & 5.14 & 7.46 & 10.64 & 13.16 \\
SO$_2$ (ppb) & 0.06 & 0.10 & 0.20 & 0.37 & 0.64 & 0.88 & 1.02 \\
CO (ppm) & 0.11 & 0.14 & 0.19 & 0.24 & 0.31 & 0.37 & 0.41 \\
PM$_{2.5}$ ($\mu$g/m$^3$) & 3.46 & 3.92 & 4.87 & 6.26 & 8.15 & 10.25 & 12.11 \\
PM$_{10}$ ($\mu$g/m$^3$) & 4.84 & 5.77 & 8.07 & 10.92 & 13.60 & 16.52 & 19.00 \\
\textbf{Meteorological} \\
Temperature ($^{\circ}$C) & 3.1 & 5.1 & 9.9 & 17.4 & 24.1 & 26.7 & 27.7 \\
Relative Humidity & 50.1 & 55.8 & 65.4 & 74.7 & 82.9 & 89.1 & 92.0 \\
Precipitation (mm) & 0.0 & 0.0 & 0.0 & 0.0 & 1.0 & 10.4 & 19.1 \\
\textbf{Socioeconomic} \\
COI Score & 6 & 11 & 28 & 45 & 66 & 77 & 83 \\
SVI Percentile & 60 & 63 & 76 & 86 & 93 & 97 & 99 \\
\midrule
Daily exacerbations & 1 & 2 & 3 & 6 & 9 & 13 & 15 \\
\bottomrule
\end{tabular}
    \caption{Percentiles for predictor variables and exacerbations over all ZCTAs during study period.}
    \label{tab:descriptive}
\end{table}

\newpage
\section{Results}

\subsection{Base models: NN and GLM}

\begin{wraptable}[8]{r}{0.5\textwidth}
    \vspace{-2.5em}
    \centering
    \begin{tabular}{c | cc}
    Lag (days) & R$^2$ ($\uparrow$) & MAE ($\downarrow$) \\
    \hline
    0 & 0.623 (0.005) & 2.10 (0.02) \\
    1 & 0.638 (0.007) & 2.06 (0.02) \\
    3 & 0.625 (0.019) & 2.09 (0.07) \\
    5 & 0.625 (0.004) & 2.077 (0.005) \\
    7 & \textbf{0.645 (0.014)} & \textbf{2.04 (0.05)} \\
    \end{tabular}
    \caption{Neural network accuracy mean (standard deviation) vs. exposure lags. $N=5$ trials. }
    \label{tab:nn_lag}
\end{wraptable}
We began by asking: over what time period do exposures influence asthma exacerbation?
Recent studies have identified associations between exacerbation rates and exposures occurring in the previous week~\citep{hsu_differential_2020,shmool_spatio-temporal_2016,huang_prediction_2025}.
However, it is less clear how interactions between variables influence subsequent outcomes.
We trained NNs to predict AE visits from measurement data averaged over the previous 1, 3, 5, and 7 days and compared to a NN trained to predict each day using instantaneous measurement data.
Performance was greater when incorporating information from previous days, and the NN achieved the lowest MAE and highest R$^2$ when averaging exposure measurements over the prior 7 days (Table~\ref{tab:nn_lag}).
The NN is a ``maximal" model whose characteristics can inform construction of simpler models.
Using its performance as a guideline, we chose to use lagged measurements averaged over the previous 7 days to fit and analyze subsequent models. 

A neural network is a black box, providing little insight into how it makes predictions~\citep{murdoch_definitions_2019}.
To obtain a more transparent model we turned to GLMs, which we hypothesized would sacrifice some predictive power.
We fit a non-interacting GLM to the same measurement and AE visit data using 7-day lagged exposures.
The GLM achieved lower predictive performance (Table~\ref{tab:accuracy}).
Despite using the same set of predictor variables, it showed a 58\% decrease in R$^2$ and an 44\% increase in MAE compared to the neural network.
However, it still captured the overall trend in exacerbations over the study period (Figure~\ref{fig:predictions}). 

\subsection{Sparse framework allows tunable complexity and accuracy}

The GLM and NN constitute two endpoints on a spectrum that trades predictive power for explainability (Figure~\ref{fig:workflow}). 
The GLM used linear relationships between exposures and log-outcomes. 
The NN generated more accurate predictions using relationships, nonlinear features, and interactions from exposures.
As a middle ground, we considered sparse dictionary models (Sparse) that select terms from a library of candidate functions.
We anticipated that Sparse models might account for greater complexity than the baseline GLM but unlike NNs, they would identify parsimonious equations and enable more transparent interpretation.

\begin{figure}[ht]
    \begin{minipage}[b]{0.5\textwidth}
        \centering
        \begin{tabular}{r | c c c}    
            Model & Params & R$^2$ ($\uparrow$) & MAE ($\downarrow$) \\
            \hline
            GLM & 28 & 0.270 (0.003) & 2.938 (0.019) \\
            Sparse & 45 (3) & 0.399 (0.004) & 2.636 (0.007) \\
            NN & 73K & \textbf{0.645 (0.014)} & \textbf{2.04 (0.05)} \\
        \end{tabular}
        \captionof{table}{Model parameter count and test accuracy (daily AE visits). Mean and standard deviation over $N=5$ trials. }
        \label{tab:accuracy}
        \vspace{1em}
        \includegraphics[width=\linewidth]{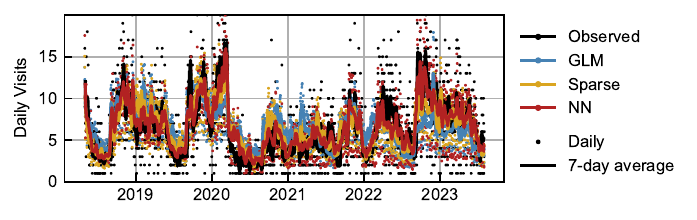}
        \captionof{figure}{Time series of average daily AE visit predictions for each model type. Solid lines show 7-day rolling averages. $N=5$ trials. }
        \label{fig:predictions}
    \end{minipage}
    \begin{minipage}[b]{0.5\textwidth}
        \centering
        \includegraphics[width=0.7\linewidth]{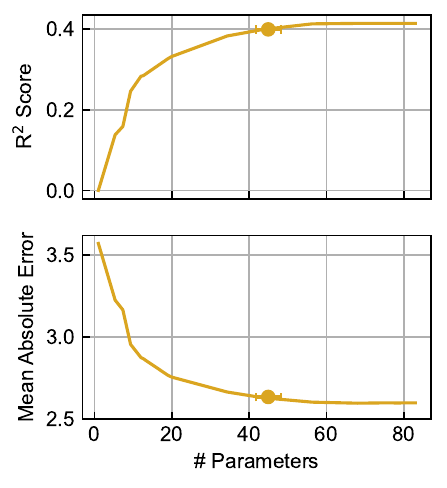}
        \captionof{figure}{Sparse model R$^2$ increases (\textit{top}) and MAE decreases (\textit{bottom}) as lower $\tau$ thresholds enable more complex models. Dot indicates model used for analysis ($\tau = 8.5$).
        }
        \label{fig:sparsity_curves}
        \end{minipage}
\end{figure}

Before evaluating our Sparse framework, we first asked: how complicated should the model be? 
The coefficient threshold parameter $\tau$ determined how many terms should be removed from the library at each algorithm iteration. 
We fit Sparse models using 25 different values of $\tau \in [0.01, 50]$. 
The test performance of each model as well as the number of terms in the model equations are summarized in Figure~\ref{fig:sparsity_curves}.
The least complex models contained fewer than 10 terms and were less accurate than the baseline GLM.
Models became more accurate as $\tau$ decreased and more terms were allowed up to a maximum of all 83 library functions at $\tau = 0.01$. 
We observed diminishing accuracy gains for models that included more than approximately 40 parameters.
For the remainder of this work, we chose a single value of $\tau = 8.5$ that selected models at the cusp of this performance plateau (Figure~\ref{fig:sparsity_curves}).
These models were more accurate than GLM, showing a 48\% increase in R$^2$ and 10\% decrease in MAE (Table~\ref{tab:accuracy} and Figure~\ref{fig:predictions}).
However, they remained below the benchmark set by the NN, suggesting that our function library which contained up to second-order polynomials (Eq.~\ref{eq:library}) could only approximate the full nonlinear NN behavior.

\subsection{Relative risk consistency across models}

\begin{figure}[t]
    \begin{minipage}[b]{0.5\textwidth}
        \centering
        \includegraphics[width=\linewidth]{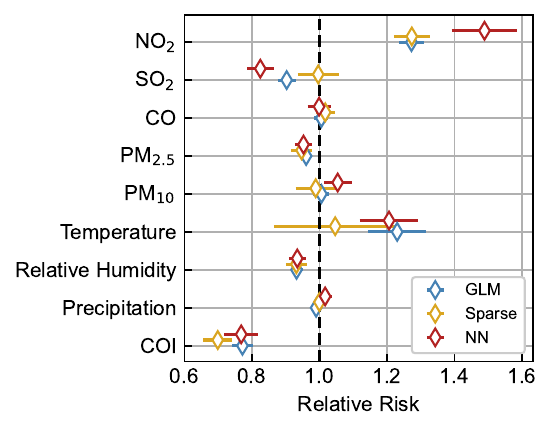}
        \captionof{figure}{Relative risks for each model. Error bars show standard deviation over $N=15$ models (5 trials of 3-fold cross validation). }
        \label{fig:risk_ratio}
    \end{minipage}
    \begin{minipage}[b]{0.5\textwidth}
        \centering
        \includegraphics[width=\linewidth]{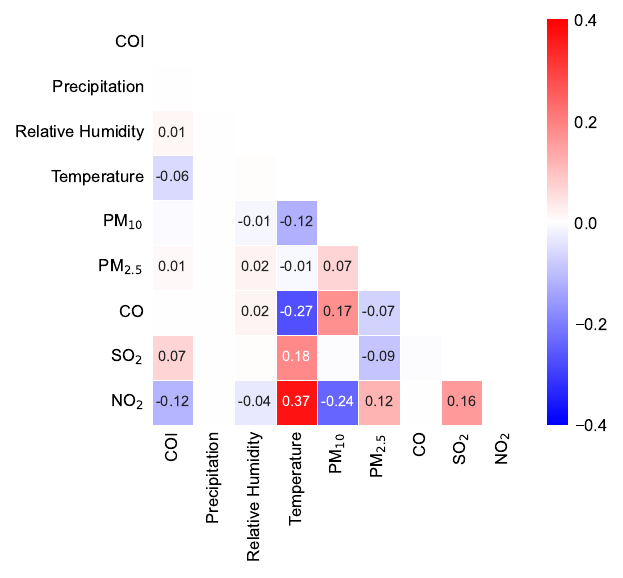}
        \caption{Relative excess risk due to interaction (RERI) for Sparse models. Average over $N=15$ models (5 trials of 3-fold cross validation) 
        }
        \label{fig:reri}
    \end{minipage}
\end{figure}

Having established how accurately each model could predict AE, we next sought to understand the relationship between a set of measurements and the corresponding model output.
We analyzed associations between exposures and predicted AE by estimating relative risks for each modeling framework. 
For a GLM, the model coefficients defined the change in log-outcomes corresponding to a one-unit (IQR) increase in each variable. For an exposure with coefficient $\beta$, the relative risk associated with an increase in that exposure was $\exp \beta$. 
For Sparse models, their mathematical form admitted direct calculation of relative risks that we could compare to the GLM results.
This calculation was more complicated, as the nonlinear interactions required accounting for the present values of all other exposures when estimating individual risks. 
The full details of how we performed this calculation are provided in the Appendix. 
For the NN models, we approximated relative risk by averaging network saliency across all predictions (see Methods). 
The results for each model and continuous exposure variable are shown in Figure~\ref{fig:risk_ratio}.

For most exposures, the associations were consistent across all modeling frameworks.
Each model identified NO$_2$ as having the strongest positive effect 
($RR_{\text{GLM}} = 1.27 \pm 0.04$, 
 $RR_{\text{NN}} = 1.49 \pm 0.10$,
 $RR_{\text{Sparse}} = 1.27 \pm 0.05 $)
and COI as having the strongest protective effect
($RR_{\text{GLM}} = 0.77 \pm 0.03$, 
 $RR_{\text{NN}} = 0.77 \pm 0.05$,
 $RR_{\text{Sparse}} = 0.70 \pm 0.04 $).
Other factors, such as CO, PM$_{2.5}$, PM$_{10}$, Relative Humidity, and Precipitation, showed similarly weaker effects for each model (see Figure~\ref{fig:risk_ratio}).

We observed discrepancies for SO$_2$ and Temperature.
While GLM and NN both identified a protective effect for SO$_2$
($RR_{\text{GLM}} = 0.90 \pm 0.03$,
 $RR_{\text{NN}} = 0.83 \pm 0.04$)
, this effect disappeared for Sparse models
($RR_{\text{Sparse}} = 0.99 \pm 0.06$).
Similarly, Temperature had a positive effect in GLM and NN trials 
($RR_{\text{GLM}} = 1.23 \pm 0.09$,
 $RR_{\text{NN}} = 1.21 \pm 0.09$)
 but not for Sparse models ($RR_{\text{Sparse}} = 1.04 \pm 0.18$).
The negative association between SO$_2$ and AE for some of our models conflicted with previous work showing harmful trends for this pollutant~\citep{guarnieri2014outdoor,greenberg_different_2016,reno_mechanisms_2015,rosser_air_2022}. 
We hypothesized that Sparse models were able to eliminate this spurious effect compared to GLM by attributing it instead to selected higher order features or interactions. 
It is unclear why the effect reappeared in NN models. 
Future studies could investigate this behavior using expanded Sparse models with broader function libraries, which could in principle approximate the more complex nonlinear features in NN models. 

\subsection{Sparse models identify interactions from data}

While Sparse models did not perform as well as NN, they were more explainable.
For example, the coefficients of selected second-order terms help define the relative excess risk due to interaction (RERI, see Methods).
This measures the change in risk due to an increase in two exposures compared to a baseline of changing each exposure independently.
Figure~\ref{fig:reri} summarizes RERI for Sparse models. 
These interactions shed some light on the relative risk discrepancies observed in Figure~\ref{fig:risk_ratio}.
For example, Sparse models identified positive interactions for SO$_2$ with NO$_2$ (RERI = 0.16), Temperature (RERI = 0.18), and COI (0.07), as well as a small negative interaction with PM$_{2.5}$ (RERI = -0.09).
Together, these interactions could have contributed to the 10\% relative risk increase for SO$_2$ compared to GLM, which eliminated the observed protective effect. 
Temperature had strong interactions with all air pollutants except PM$_{2.5}$ and showed a weaker interaction with COI. 
When aggregating the corresponding relative risk over the heterogeneous study area and timespan (see Appendix for details), this may have reduced its average effect size compared to the GLM baseline but also increased its variance (Figure~\ref{fig:risk_ratio}).

%% file: chapters/discussion.tex
\section{Discussion}

Here, we compared distinct approaches which trade predictive power for explainability to model pediatric asthma exacerbation in southeast Virginia.
Generalized linear models have been applied to similar problems and datasets to assess individual impacts of environmental and pollution factors~\citep{deflorio-barker_cardiopulmonary_2019,james_health_2018,tinling_repeating_2016,kiser_particulate_2020}. 
While these may in principle capture targeted interactions, it is difficult to extend them to datasets containing a wide range of potential exposures over time.
Deep neural networks can make stronger predictions but are challenging to interpret and explain~\citep{murdoch_definitions_2019}.
In this study, deep neural networks were a target, representing an upper bound on a model's ability to describe the dataset.
To bridge the gap between GLM and NN models, we introduced a new framework based on sparse dictionary learning that allows tunable model complexity and automatic identification of higher-order features and interactions.

The sparsity objective required the model to select a minimal number of terms from a broad library, which aided interpretability.
Other methods to examine interactions include choosing to include such a term \textit{a priori} or conducting stratified analyses.
This is challenging with complex multi-modal datasets where the number of potential interactions grows rapidly.
The strongest interaction we found was between NO$_2$ and Temperature (Figure~\ref{fig:reri}).
We hypothesize that this may be related to ozone (O$_3$), which has been associated with pediatric asthma severity~\citep{kelchtermans_ambient_2024} but was omitted from our analysis as it was not recorded consistently during the study period.
Fitting a GLM on days when O$_3$ measurements were available showed a positive effect for O$_3$ and a reduced risk associated with temperature (Figure~\ref{fig:ozone}).
Studies have shown that higher temperatures enhance photochemical activity and alter NO$_x$ cycling, influencing NO$_2$ concentrations through changes in photolysis rates, ozone production, and temperature-sensitive emissions processes~\citep{seinfeld2016atmospheric,bloomer2009observed,he2013high, golbazi2023impacts}. However, these relationships are highly nonlinear and depend on local photochemical regimes, as ozone formation sensitivity varies between NO$_x$-limited and volatile organic compound (VOC) limited environments commonly associated with rural and urban regions, respectively~\citep{seinfeld2016atmospheric}. In addition, elevated temperatures have been associated with increased anthropogenic NO$_x$ emissions due to higher electricity demand and energy production during heat events~\citep{abel2017response,he2013high}, along with enhanced biogenic and soil-related NO$_x$ contributions under extreme temperature conditions~\citep{romer2018effects,goldberg2021tropomi}.

Socioeconomic indicators showed consistent effects across all models.
Higher child opportunity was associated with lower risk of asthma exacerbation and models were consistent when substituting COI for SVI (note that higher social vulnerability, or higher disadvantage, was associated with a higher risk, see Figure~\ref{fig:svi_sensitivity}). 
Interactions identified with COI include NO$_2$ (negative), Temperature (negative) and SO$_2$ (positive).
This may be intertwined with the coarse localization of pollutant measurements and exposures.
In our study area, neighborhoods with low COI tend to be closer to cities, highways, and dense roads (Figure~\ref{fig:coi_map} left).
Here, high traffic activity is a major source of NO$_2$ that could potentially elevate pollutant concentration in these lower-COI communities.
In contrast, higher-COI neighborhoods tend to be near the coast, where marine vessel emissions may contribute to higher local SO$_2$ concentrations (Figure~\ref{fig:coi_map} right)~\citep{golbazi2023impacts}. 
This suggests that the sparse models may identify and apply COI-pollutant interactions to compensate for limited ambient air quality resolution.
The interaction between COI and temperature may be explained by the urban heat island effect, which disproportionately affects socioeconomically disadvantaged populations and urban communities in this region~\citep{golbazi2025high}.
Future work integrating our modeling frameworks with locally-calibrated atmospheric chemistry models may shed light on this behavior.

Neighborhood level socioeconomic indicators and subcomponents of COI/SVI may also be useful in estimating levels of individual exposure to ambient air quality during acute events such as wildfire smoke.
The latter may modulate the composition of atmospheric particulate matter to impact AE ~\citep{reid_differential_2016,wang_differential_2025} and also transport infectious bacterial and fungal cells~\citep{kobziar_wildfire_2020,moore_wildland_2021}. 
A preliminary analysis in our study area shows the presence of medium to heavy atmospheric smoke enhances AE risk for PM$_{10}$ (Figure~\ref{fig:stratify_smoke}). 
This analysis, which was limited by the small number of heavy smoke days during the observation period, also found smoke-mediated protective effects for CO and PM$_{2.5}$ which may require more detailed follow-up to understand. 
Atmospheric chemistry modeling to identify how elevated smoke presence contributes to ground level ambient air quality and pollutant concentration could help in future analyses on AE during extreme events.

The correspondence between NN saliency and GLM risk ratio suggests the former could be a reasonable metric of interpretation in future work.
The neural network achieved higher performance than either GLM or Sparse models, which we attributed to its ability to construct complex nonlinear features.
Sparse models shed some light on these features but did not fully explain or match NN predictions.
Expanded libraries featuring higher order polynomials or other nonlinear functions may bridge the performance gap.
Alternatively, integrating sparse dictionary learning within NN training~\citep{champion_data-driven_2019} could produce more transparent features to enhance explainability without sacrificing performance.
Extending these data-driven methods to learn interactions between population groups~\citep{tkachenko_stochastic_2021,seara_sociohydrodynamics_2025} may improve understanding of how social and familial conditions modulate asthma severity~\citep{wood_family_2007,alcala_longitudinal_2023}.
Our approach here, which blended \textit{post hoc} NN analysis with restricted Sparse models with inherent explainability, demonstrates an alternative path to characterize syndemic interactions influencing asthma exacerbation.
Linking biostatistics and deep learning in future work may aid prediction and spatio-temporal analysis in environmental public health studies. 

%% file: chapters/limitations.tex
\section{Limitations of the study}

Our study had several limitations.
Our data protocol did not track patients over time and thus could not distinguish between primary and secondary exacerbations. 
We also could not exclude that some AE visits were follow-up visits either for an acute event or ongoing maintenance.
Using a specific code such as status asthmaticus would likely miss many cases of AE.
Our analysis also relies on accurate patient coding which could not be confirmed without patient identifiers. 
Air quality measurements at federal reference monitors may not provide sufficient granularity to capture individual-level pollutant exposures. 
In addition, they do not necessarily indicate indoor air quality.  
While we hypothesize that interactions with socioeconomic indicators may have partially accounted for this limited localization, we hope to improve our exposure estimates in future work using atmospheric chemistry models coupled to mesoscale measures of the local built environment. 
We note that our de-identified dataset would limit assessment at resolutions finer than the zip code level. 
Here, we predicted daily exacerbations as a static problem, but populations are fundamentally dynamic. 
Future studies could adapt our framework to examine correspondences between autoregressive time series models, recurrent neural networks, and frameworks for sparse identification of nonlinear dynamical systems. 

%% file: chapters/appendix.tex
\section{Sparse Dictionary Modeling}

Sparse dictionary modeling aims to model a target variable using a linear combination of nonlinear terms from a candidate function library $L$. 
Here, we performed Poisson regression on the AE visits $y$, so the objective was to find an equation that estimates $\log y$. 
We first assembled a column vector whose rows were visit counts at each ZCTA and day.
Next, we used a preset function library to construct a matrix whose columns represented each possible mathematical term and whose rows gave the value of that term at each site and day. 
For a simplified problem with one location, two inputs $x_1, x_2$, and a function library $\{ x, x^2, x \times y \}$, the goal would be to find weights $\beta$ that optimize
\begin{small}
\begin{equation}
    \log \begin{bmatrix}
        y(5/1/18 ) \\
        y(5/2/18 ) \\
        y(5/3/18 ) \\
        \vdots
    \end{bmatrix}
    = 
    \begin{bmatrix}
        x_1 (5/1/18) & x_2(5/1/18) & x_1^2 (5/1/18) & x_2^2 (5/1/18) & x_1 \times x_2 (5/1/18) \\
        x_1 (5/2/18) & x_2(5/2/18) & x_1^2 (5/2/18) & x_2^2 (5/2/18) & x_1 \times x_2 (5/2/18) \\
        x_1 (5/3/18) & x_2(5/3/18) & x_1^2 (5/3/18) & x_2^2 (5/3/18) & x_1 \times x_2 (5/3/18) \\
        \vdots & \vdots & \vdots & \vdots & \vdots 
    \end{bmatrix}
    \begin{bmatrix}
        \beta_1 \\
        \beta_2 \\
        \beta_3 \\
        \beta_4 \\ 
        \beta_5 
    \end{bmatrix}
    \label{eq:sparse_example}
\end{equation}
\end{small}
For this work, our function library (Equation~\ref{eq:library}) contained all second-order polynomial terms for continuous variables concatenated with the binary indicators for weekday and month. This produced a total of 83 possible mathematical terms that could enter the model. 
To control the complexity of the model and limit the number of terms that could contribute to predictions, we used the sequential thresholding approach introduced in~\citep{brunton_discovering_2016}. 
At each optimization iteration, any terms whose coefficient $\beta$ is below a threshold $\tau$ is removed from the library during subsequent iterations. 
The fitting algorithm is represented below.

\begin{minipage}{0.6\textwidth}
\begin{algorithm}[H]
\caption{Sparse Dictionary Regression with Sequential Thresholding}
\label{alg:spr}
\SetAlgoLined
Initialize exposures $\mathbf{x}$, indicators $\mathbf{u}$, outcomes $y$ \\
Initialize library function $L$ \\
Initialize threshold $\tau$ and mask $\mathbf{m} = \mathbf{1}$ \\
\For{$t$ in  $1\dots N$}{
    $\bm{\beta}[\mathbf{m}] \leftarrow$ {Poisson Regression on} $(L(\mathbf{x}, \mathbf{u})[\mathbf{m}], y)$ \\
    $\mathbf{m} \leftarrow \bm{\beta} \ge \tau$ \\
    $\beta[\neg \mathbf{m}] \leftarrow \mathbf{0} $
}
\end{algorithm}
\end{minipage}

\subsection{Relative Risk Calculation}

\newcommand{\at}[2][]{#1|_{#2}}

The relative risk is the proportional change in population risk caused by a one-unit increase in a specified exposure. For a GLM with a Poisson distribution and log-link function, exponentiating model coefficients provides an estimate of relative risk. However, the sparse models included interactions and thus the effect of changing one exposure level depended on the present values of all other exposures. Thus, we estimated model relative risk by aggregating baseline and exposed predicted counts across the entire dataset as described below.

First, we derive the mean and variance of a function of random variables $f(\beta_1, \beta_2, \dots)$. Suppose these random variables follow a joint distribution $\beta_i \sim \mathcal{N}( \mu_i, \Sigma_{ij})$. For any specific values of $\beta_i$, we can approximate $f$ using a Taylor series expansion
\begin{align}
   f(\beta_i)  &= f(\mu_i) + 
   \sum_i \frac{\partial f}{\partial \beta_i} \at[\bigg]{\beta=\mu} (\beta_i - \mu_i) + 
   \frac{1}{2} \sum_{ij} \frac{\partial^2 f}{\partial \beta_i \partial \beta_j} \at[\bigg]{\beta = \mu }
   (\beta_i - \mu_i)(\beta_j - \mu_j) + O\left[ (\beta - \mu)^3 \right] 
\end{align}
We can compute the mean of this function by taking
\begin{align}
    \mathbb{E} \left[ f \right] &= f(\mu_i) + 
   \sum_i \frac{\partial f}{\partial \beta_i} \at[\bigg]{\beta=\mu} \mathbb{E} \left[ \beta_i - \mu_i \right] + 
   \frac{1}{2} \sum_{ij} \frac{\partial^2 f}{\partial \beta_i \partial \beta_j} \at[\bigg]{\beta=\mu}
   \mathbb{E} \left[ (\beta_i - \mu_i)(\beta_j - \mu_j) \right] + \dots \\
   &= f(\mu_i) + \frac{1}{2} \sum_{ij} \frac{\partial^2 f}{\partial \beta_i \partial \beta_j} \at[\bigg]{\beta=\mu} \Sigma_{ij} 
\end{align}
Similarly, we can compute the variance as $\mathbb{E} \left[ f^2 \right] - \mathbb{E}\left[f\right]^2$. Here, we have
\begin{align}
    \mathbb{E} \left[ f^2 \right] &= \mathbb{E} \left[ 
        \left( f(\mu_i) + \sum_i \frac{\partial f}{\partial \beta_i}(\beta_i -\mu_i) + \frac{1}{2} \sum_{ij} \frac{\partial^2 f}{\partial \beta_i \partial \beta_j} (\beta_i-\mu_i)(\beta_j-\mu_j) \right)^2 \right] \\
        &= f(\mu_i)^2 + \sum_{ij} \left( \frac{\partial f}{\partial \beta_i} \frac{\partial f}{\partial \beta_j} + f(\mu) \frac{\partial^2 f}{\partial \beta_i \partial \beta_j} \right) \Sigma_{ij} + \dots 
\end{align}
Thus, we have that
\begin{align}
    \sigma^2_f &= \sum_{ij} \frac{\partial f}{\partial \beta_i} \frac{\partial f}{\partial \beta_j} \Sigma_{ij}
    \label{eq:var_rr_delta}
\end{align}
For Sparse models, we obtained $\mu_i, \Sigma_{ij}$ from the final \texttt{statsmodels} fit result for all non-zero selected coefficients and applied the following definition for relative risk:
\begin{align}
    RR &= \frac{\sum_{t=1}^T \sum_{\ell=1}^L \exp \beta_k w_k(t, \ell) }{\sum_{t=1}^T \sum_{\ell=1}^L \exp \beta_k u_k (t, \ell) } = \frac{E}{B}
\end{align}
Here $t$ runs over all time points, $\ell$ runs over all spatial regions, $u_k (t, \ell)$ is the value of a feature in the candidate library, and $w_k (t, \ell)$ is the value of the same feature after effecting an exposure increase $x_i (t, \ell)  \rightarrow x_i (t, \ell) + 1$. 
Thus, this defines relative risk as the relative increase in \emph{total} event rate across the study period in response to a uniform one-unit (interquartile range) increase in an exposure.
Using this, we computed the derivatives (adopting the shorthand $\partial_i f \equiv \partial f / \partial \beta_i$)
\begin{align}
    \partial_i RR &= \frac{B \,\partial_i E - E \, \partial_i B
    }{B^2}
\end{align}
\begin{align}
    \partial_{ij} 
    &= \frac{
        B^3 \, \partial_{ij} E - E\, B^2\, \partial_{ij} B - B^2 \, \left( \partial_i E \, \partial_j B + \partial_i B \, \partial_j E \right) + 2 \, E \, B \, \partial_i B \, \partial_j B
    }{B^4}
\end{align}
From the definitions of $E, B$, we have
\begin{align}
    \partial_i E &= \sum_{t,\ell} w_i(t, \ell) \cdot  \exp \beta_k w_k (t, \ell) \\
    \partial_{ij} E &= \sum_{t,\ell} w_i(t ,\ell) w_j (t, \ell) \exp \beta_k w_k (t, \ell)
\end{align}
and identical definitions for $\partial_i B,\ \partial_{ij} B$ with $u \leftrightarrow w$. 
We used these to compute $\partial_i RR,\ \partial_{ij} RR$ and estimate Eq.~\ref{eq:var_rr_delta} as an uncertainty on each relative risk.
We also verified that this procedure produces the correct values for GLM models to ensure consistency. 

\newpage 
\begin{figure}
    \begin{minipage}[b]{0.5\textwidth}
        \centering
        \includegraphics[width=\linewidth]{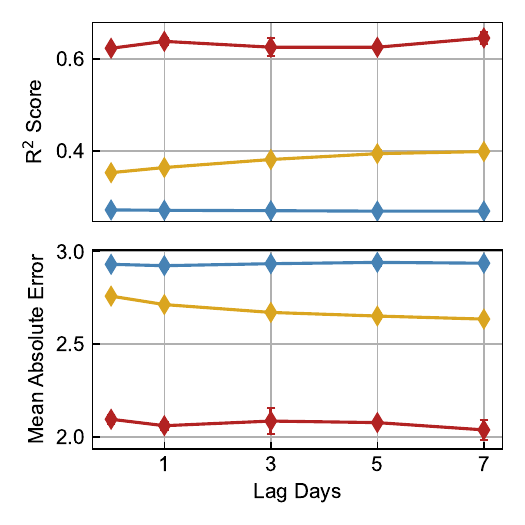}
        \captionof{figure}{R$^2$ (\textit{top}) and MAE (\textit{bottom}) vs. exposure lags for each model type. }
        \label{fig:lag_accuracy}
    \end{minipage}
    \begin{minipage}[b]{0.5\textwidth}
        \centering
        \includegraphics[width=\linewidth]{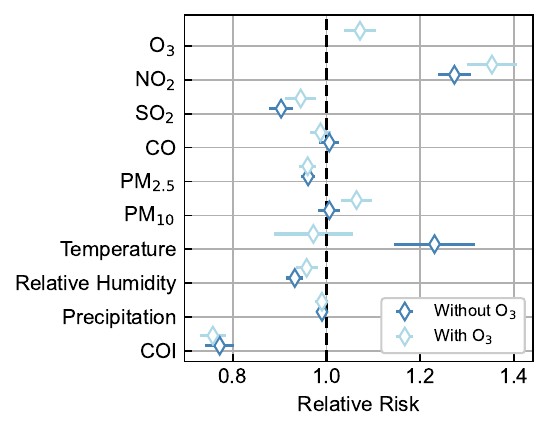}
        \captionof{figure}{Comparing GLM relative risks with and without O$_3$ as a model input.}
        \label{fig:ozone}
    \end{minipage}\end{figure}

\begin{figure}
    \centering
    \includegraphics[width=\linewidth]{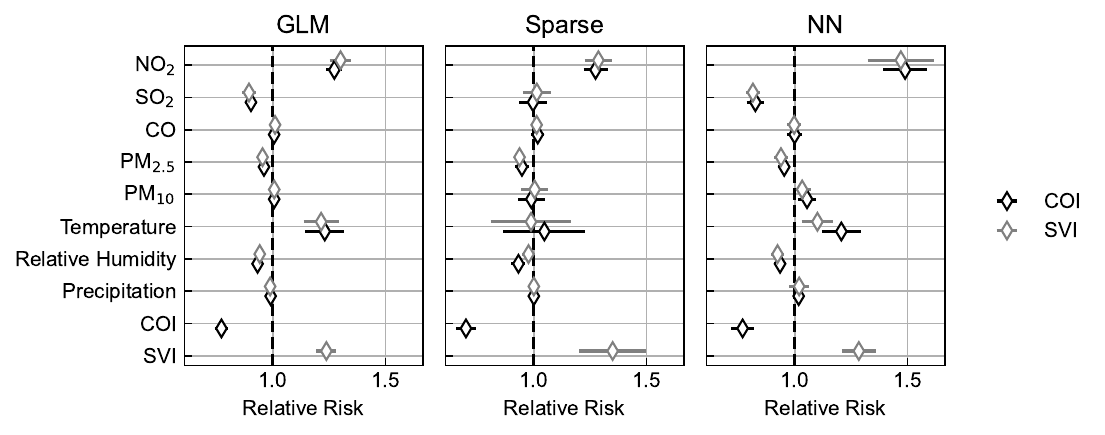}
    \caption{Models' relative risks are consistent when using SVI (grey) instead of COI (black).
        Relative risks for COI and SVI are opposite as disadvantaged neighborhoods show low COI and high SVI. 
    }
    \label{fig:svi_sensitivity}
\end{figure}

\begin{figure}
    \centering
    \includegraphics[width=\linewidth]{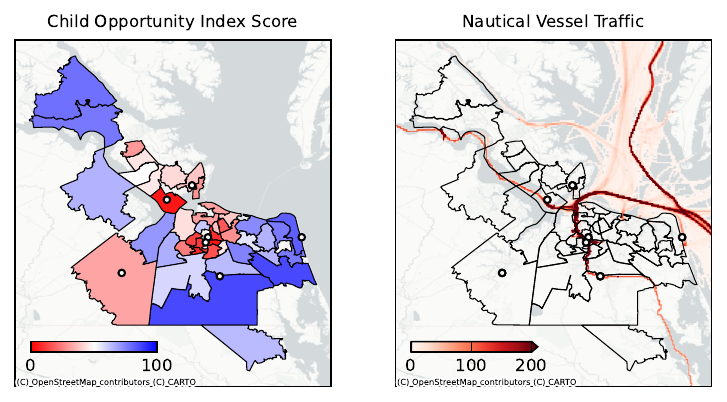}
    \caption{
        (\textit{Left}) Child opportunity index percentiles for each ZCTA in the study area (averaged 2018-2023). 
        (\textit{Right}) Nautical vessel traffic through the study area for a representative month (May 2018). Data from U.S. Vessel Traffic Automatic Identification System~\citep{us_vessel_traffic}.
        Black circles mark Hampton Roads area cities}
    \label{fig:coi_map}
\end{figure}

\begin{figure}
    \centering
    \includegraphics[width=0.9\linewidth]{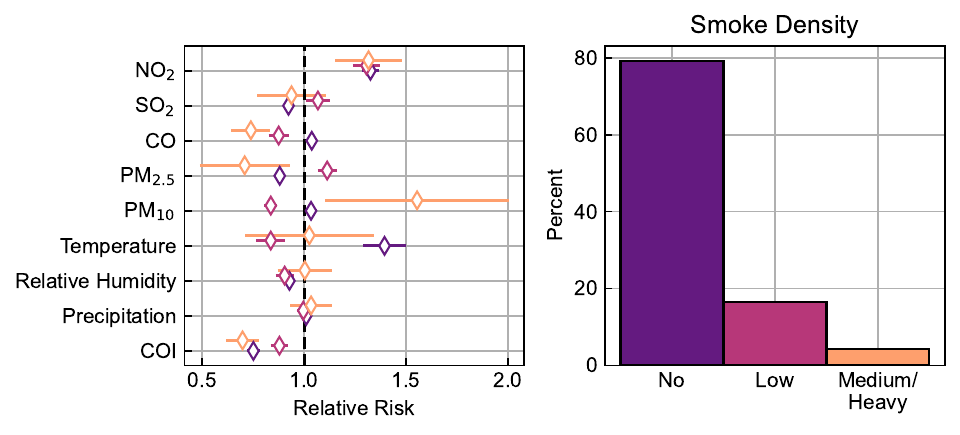}
    \caption{GLM relative risks stratified by presence of wildfire smoke. Smoke data from NOAA Hazard Mapping System~\citep{noaa_hms_nodate}}
    \label{fig:stratify_smoke}
\end{figure}